%% file: neurips_2026.tex
\documentclass{article}

 \usepackage[preprint]{neurips_2026}

\input{custom}

\title{Learning to Correct Geometry in Generated Videos}

\author{
Weijie Lyu$^{1}$ \quad
Tiancheng Shen$^{1}$ \quad
Jason Li$^{2}$ \quad
Yujing Wang$^{3}$ \\
\textbf{Xiaoyuan Wang}$^{4}$ \quad
\textbf{Haobo Yuan}$^{1}$ \quad
\textbf{Yizhou Zhao}$^{4}$ \quad
\textbf{Ming-Hsuan Yang}$^{1}$ \\[1em]
$^{1}$University of California, Merced \qquad
$^{2}$Nanyang Technological University \\
$^{3}$Peking University \qquad
$^{4}$Carnegie Mellon University
}

\newcommand{\ourmethod}{\textit{VGC}}
\newcommand{\ourdataset}{\textit{VG-Pairs}}

\begin{document}

\maketitle

\vspace{-8mm}
\begin{figure}[h!]
    \centering
\includegraphics[width=\textwidth]{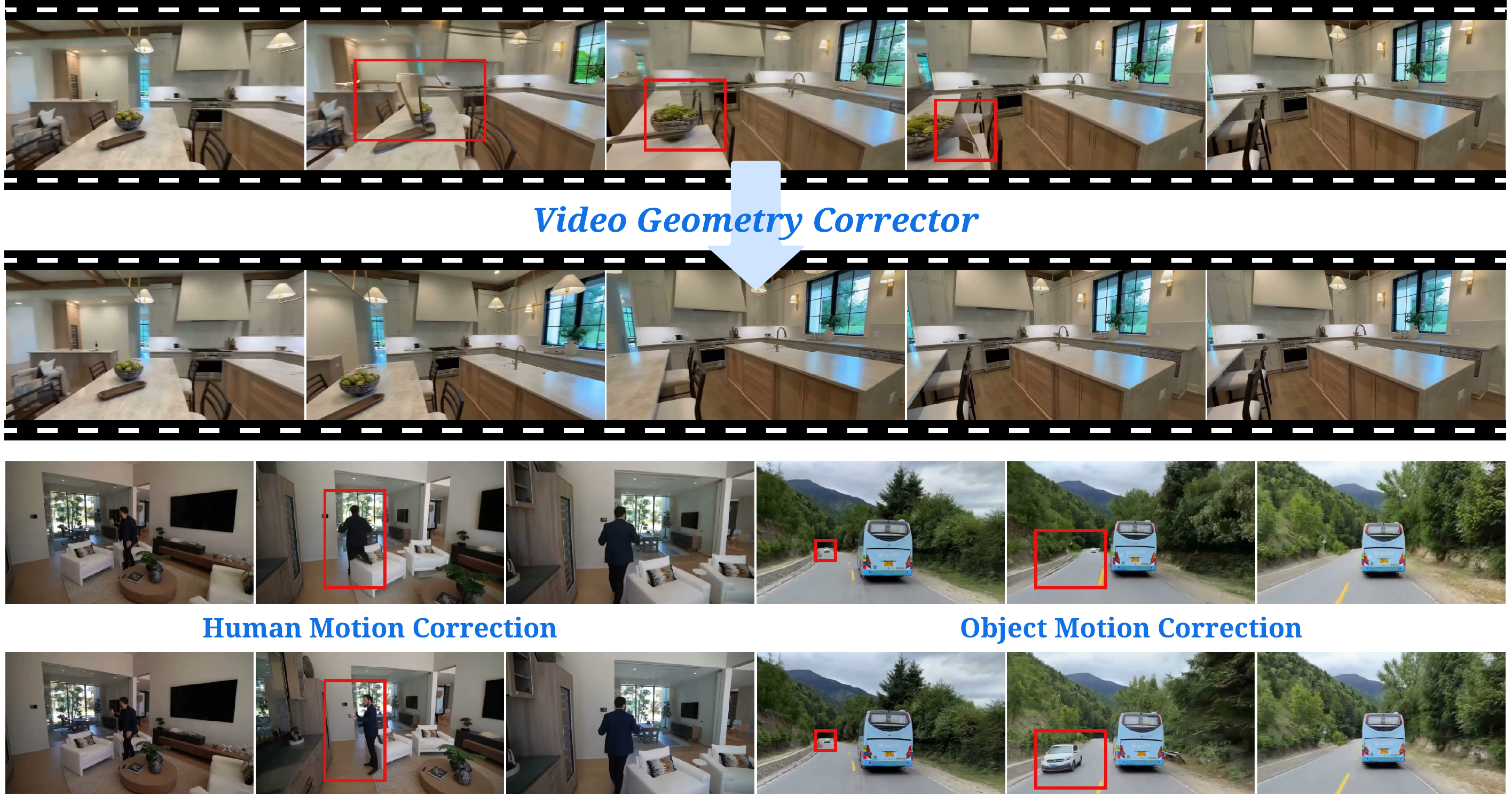}
\vspace{-4mm}
\caption{\textbf{Video Geometry Corrector} ({\ourmethod}) corrects geometry distortions in generated videos. Under large camera motion, modern video generators often produce warped structures, inconsistent parallax, and drifting scene layouts (red boxes). {\ourmethod} refines these outputs into geometry-consistent videos. When scaled to a larger backbone, it further extends to general video refinement, including dynamic scenes with distorted human or object motion. 
Project page: \url{https://weijielyu.github.io/VGC}.}
\label{fig:teasor}
\end{figure}

\input{sec/0_abstract}
\input{sec/1_intro}

\input{sec/2_related_work}

\input{sec/3_method}
\input{sec/4_exp}

\input{sec/5_conclusion}
\input{sec/7_ack}

{
    \bibliographystyle{unsrt}
    \bibliography{main}
}

\end{document}

%% file: custom.tex
\usepackage[utf8]{inputenc} 
\usepackage[T1]{fontenc}    
\usepackage[pagebackref=true,breaklinks=true,colorlinks,citecolor=blue,urlcolor=blue,linkcolor=blue,bookmarks=false]{hyperref}
\usepackage{url}            
\usepackage{booktabs}       
\usepackage{amsfonts}       
\usepackage{nicefrac}       
\usepackage{microtype}      
\usepackage{xcolor}         
\usepackage[table]{xcolor}
\usepackage{color, colortbl}
\usepackage{amsmath}
\usepackage{graphicx}
\usepackage{subcaption}
\usepackage{tabu}
\usepackage{tikz}
\usepackage{array}
\usepackage{adjustbox}
\usepackage[font=footnotesize,labelfont=footnotesize]{subcaption}
\usepackage[font=small,labelfont=footnotesize]{caption}
\usepackage{pifont}
\usepackage{placeins}
\usepackage{paralist}
\usepackage{enumitem}
\usepackage{multirow}

\definecolor{Gray}{gray}{0.9}
\definecolor{Celadon}{rgb}{0.67, 0.88, 0.69}
\definecolor{Cream}{rgb}{1.0, 0.99, 0.82}
\definecolor{DarkGreen}{RGB}{21,100,52}
\definecolor{DarkRed}{RGB}{139,0,0}
\definecolor{DarkRed2}{RGB}{180,0,0}
\definecolor{MyGreen}{rgb}{0, 0.55, 0}
\definecolor{LightYellow}{RGB}{255, 255, 204}
\definecolor{LightGreen}{rgb}{0.80, 1.0, 0.80}

\newcolumntype{M}[1]{>{\centering\arraybackslash}m{\#1}}

\usepackage{amssymb}

\usepackage{xcolor}
\definecolor{darkgreen}{RGB}{0,120,0}
\definecolor{darkred}{RGB}{180,0,0}

\newcommand{\metricgain}[2]{\makebox[3.0em][r]{#1}\,{\color{darkgreen}(#2)}}
\newcommand{\metricloss}[2]{\makebox[3.0em][r]{#1}\,{\color{darkred}(#2)}}

\usepackage{booktabs}
\usepackage{multirow}
\usepackage{makecell}
\usepackage{xcolor}
\usepackage{graphicx}
\setitemize{noitemsep,topsep=0pt,parsep=0pt,partopsep=0pt}

\usepackage{algorithm}
\usepackage{algpseudocode}

\usepackage{caption}

%% file: sec/0_abstract.tex
\begin{abstract}
\vspace{-2mm}
Generated videos often lose geometric consistency under camera motion: structures warp under viewpoint change, parallax becomes inconsistent, and smooth camera movement may collapse into shot-like transitions.
We argue that this failure is largely a data problem.
In web-scale video corpora, most visible motion comes from dynamic objects captured by static or weakly moving cameras, leaving only sparse supervision for viewpoint-induced appearance changes.
To address this, we propose Video Geometry Corrector ({\ourmethod}), a post-hoc video-to-video model trained on synthetic paired supervision.
Starting from real videos with diverse camera motion and pose annotations, we construct {\ourdataset} with a synthetic paired-supervision data generation pipeline.
Each video clip is treated as a geometry-consistent target and paired with a distorted counterpart synthesized by a pretrained image-to-video inpainting model using trajectory-adaptive keyframes.
{\ourmethod} learns to map the distorted input to the geometry-consistent target without modifying the source generator, and uses optical flow as an auxiliary cue to help disentangle camera motion from object motion.
Experiments on pose-conditioned and prompt-driven benchmarks show that {\ourmethod} improves geometric consistency across multiple source generators, and scaling to a larger backbone further extends the framework toward more general video refinement.
%
\end{abstract}

%% file: sec/1_intro.tex
\section{Introduction}
Recent video generation models can synthesize visually compelling clips from text, images, or video conditions~\cite{sora,congvideox,hunyuanvideo,wan,seedance,wu2024motionbooth,upscaleavideo,zhang2025transform}.
However, their apparent ability to behave as world simulators often breaks down once the camera moves.
Rigid structures may warp under viewpoint changes, parallax may become inconsistent, scene layout may drift across frames, and continuous camera motion may collapse into shot-like transitions.
These failures are especially severe under large viewpoint changes, showing that current models often generate plausible frame sequences but fail to maintain a stable 3D scene layout.
We argue that camera-motion-induced geometric distortion is largely a data problem.
In web-scale video corpora, most visible motion comes from dynamic objects captured by static or weakly moving cameras, leaving only sparse supervision for viewpoint-induced appearance changes.
As a result, camera motion and object motion are easily entangled: when asked to generate a moving-camera shot, a model may explain the visual change by deforming scene structure rather than maintaining a stable 3D layout.

To address this problem, existing works have largely followed two directions.
The first makes the generator more geometry-aware by introducing 3D supervision or spatial priors during training~\cite{geovideo,geometryforcing,epipolar,world,gen3c}.
However, as foundation models evolve rapidly and are often accessible only through checkpoints or inference APIs, modifying the source model is not always feasible.
The second direction applies post-hoc video refinement~\cite{upscaleavideo,seedvr,venhancer}.
These methods primarily target pixel-level restoration, such as super-resolution, denoising, and temporal smoothing, thereby improving appearance without explicitly correcting the underlying geometric distortion.
Post-hoc geometry correction, which requires reasoning about 3D structure under camera motion, remains comparatively underexplored.

The key bottleneck is the training data.
Training a geometry corrector requires paired supervision: a geometry-distorted video and a geometry-consistent target that follows the same camera trajectory.
Such pairs are \textit{nearly impossible} to collect in the real world.
%
%
To address this challenge, we construct the {\ourdataset} dataset using a paired-supervision data generation pipeline.
Starting from real videos with diverse camera motion and annotated poses, we synthesize distorted counterparts using Wan 2.1-Fun-InP, a pretrained image-to-video inpainting model based on Wan~\cite{wan}.
Instead of conditioning only on the first and last frames, we adopt a trajectory-adaptive keyframe strategy that inserts additional keyframes at abrupt pose changes, enabling stable generation of shorter subsegments that are later stitched together.
This process produces paired videos with identical semantics and similar camera motion, but different geometric fidelity.

Built on {\ourdataset}, we propose Video Geometry Corrector ({\ourmethod}), a video-to-video diffusion model that transforms geometry-distorted videos into structurally coherent outputs.
{\ourmethod} restores stable scene geometry without regenerating videos from scratch, preserving scene content and object appearance.
To further disentangle camera motion from object motion, we incorporate optical flow as an auxiliary structural cue.
Optical flow separates globally coherent camera-induced motion from local object dynamics, helping the model determine what to correct and what to preserve.

Although our main focus is correcting geometry in static scenes under camera motion, scaling {\ourmethod} to larger-capacity models naturally extends its capability to dynamic videos containing human and object motion, moving toward general video refinement.
This flexibility distinguishes {\ourmethod} from prior geometry-focused approaches~\cite{videogpa,geometryforcing}, which typically focus on static environments.
The main contributions of this work are:
\begin{itemize}[leftmargin=6mm, labelsep=0.5em, nosep]
\item We identify a critical gap in post-hoc video refinement and attribute camera-induced geometric distortions to insufficient supervision for viewpoint-induced appearance changes.
\item We address the paired-data bottleneck with {\ourdataset}, a synthetic paired-supervision dataset built from real videos with camera motion and their trajectory-aware geometric distortions.
\item We propose {\ourmethod}, a post-hoc video-to-video geometry corrector augmented with optical flow cues, demonstrating strong structural restoration, cross-generator transferability, and scalability to dynamic videos.
\end{itemize}

%% file: sec/2_related_work.tex
\section{Related Work}
\vspace{-2mm}
\textbf{Geometry-consistent video generation.}
Recent video diffusion models such as Sora~\cite{sora}, CogVideoX~\cite{congvideox}, HunyuanVideo~\cite{hunyuanvideo}, and Wan~\cite{wan} achieve strong visual quality, yet maintaining geometric consistency under camera motion remains challenging.
Most existing approaches address this issue at the generator or training level.
Some methods introduce explicit geometric supervision during training.
GeoVideo~\cite{geovideo} incorporates depth prediction and cross-frame consistency losses.
Geometry Forcing~\cite{geometryforcing} aligns intermediate features with pretrained 3D foundation models~\cite{vggt}, and Epipolar Geometry Improves Video Generation Models~\cite{epipolar} introduces epipolar constraints during fine-tuning.
Other approaches improve geometry through post-training alignment or reward optimization.
VideoGPA~\cite{videogpa} derives preference signals from reprojection consistency, VideoREPA~\cite{videorepa} transfers physical reasoning from video foundation models, and VGGRPO~\cite{vggrpo} extends geometry-aware rewards into latent space for dynamic scenes.
A third direction incorporates explicit 3D structure into the generation pipeline.
World-consistent Video Diffusion~\cite{world} maintains a point-cloud memory for spatial consistency, while GEN3C~\cite{gen3c} uses geometric priors for camera-controlled video generation.
In contrast, {\ourmethod} frames geometry correction as a post-hoc video-to-video task, avoiding modification of the source generator and enabling transfer across model families.

\textbf{Video refinement and enhancement.}
A separate line of work improves generated or degraded videos through post-processing.
Upscale-A-Video~\cite{upscaleavideo}, SeedVR~\cite{seedvr}, and SeedVR2~\cite{seedvr2} focus on restoration and super-resolution, progressing from multi-step diffusion upscaling to single-step arbitrary-resolution enhancement.
Meanwhile, several works~\cite{zhang2025transform,xue2025ultravideo} introduce new datasets to refine high-resolution video generation.
VEnhancer~\cite{venhancer} jointly improves spatial and temporal quality using a space-time controller on pretrained diffusion priors.
Enhance-A-Video~\cite{enhanceavideo} enhances temporal consistency in DiT-based models without retraining.
ReVision~\cite{revision} refines coarse videos using 3D object-centric representations and physical motion priors.
Despite improving visual quality, temporal smoothness, and motion plausibility, these methods largely overlook post-hoc correction of scene-level geometric consistency under camera motion.
{\ourmethod} differs by targeting geometry correction, using paired supervision constructed from real videos to repair structural inconsistency while preserving scene content and motion.

%% file: sec/3_method.tex
\vspace{-2mm}
\section{Method}
\vspace{-2mm}
Our goal is to restore geometric consistency in generated videos under camera motion.
We attribute these failures to insufficient supervision for viewpoint-induced appearance change in existing video data.
Instead of modifying each video generation model to learn stronger geometric priors, we formulate geometry correction as a post-hoc video-to-video task and train a separate corrector with paired supervision.
This decouples geometry correction from video generator design, making our method applicable across different model families.

Our framework consists of two stages.
First, we build {\ourdataset} by synthesizing paired distorted inputs and geometry-consistent targets from real videos with diverse camera motion and pose annotations.
Second, we train a video-to-video diffusion model, {\ourmethod}, on these pairs to correct geometry.
Optical flow extracted from the distorted input serves as an auxiliary cue for disentangling camera motion from scene dynamics.

\input{fig_text/keyframes}
\vspace{-2mm}
\subsection{Paired Supervision from Videos with Camera Motion}
\vspace{-2mm}
\label{sec:data}
The central design choice is to use videos with diverse camera motion not merely as additional training data for a generator, but as a source of correction pairs for a video-to-video task.
Hence, we construct {\ourdataset}, a dataset of geometry-distorted videos paired with geometry-consistent supervision, using the following data generation pipeline.
Let $\mathbf{x} = (x_1, \ldots, x_T)$ denote a real video clip with camera poses $\mathbf{p} = (p_1, \ldots, p_T)$, where each pose is represented as $p_i = (\mathbf{t}_i, \mathbf{q}_i)$ with camera position $\mathbf{t}_i \in \mathbb{R}^3$ and rotation $\mathbf{q}_i \in S^3$.
We treat $\mathbf{x}$ as the geometry-consistent target and synthesize a distorted counterpart, $\mathbf{y}$, using Wan 2.1-Fun-InP, a pretrained image-to-video inpainting model based on Wan~\cite{wan}.
Given a start frame and an end frame, this model generates the intermediate video that connects them.

To make $\mathbf{y}$ useful for supervision, it should preserve the scene identity and approximately follow the camera trajectory of $\mathbf{x}$ while still exhibiting geometric errors.
We therefore select a trajectory-adaptive set of keyframes $K = \{k_1, \ldots, k_{m+1}\}$ with $1=k_1<\cdots<k_{m+1}=T$, and generate one video segment between each consecutive keyframe pair.
Figure~\ref{fig:keyframes} illustrates the motivation: for clips with large, nonlinear camera motion, conditioning only on the first and last frames can lead the inpainting model to produce a shortcut trajectory that deviates from the ground-truth path, whereas adaptive keyframe selection yields distorted videos that remain close to the target trajectory.

We first estimate the total amount of camera motion in the clip.
For each adjacent frame pair, we measure rotational change by the geodesic distance between quaternions, $\Delta\phi_i = 2\arccos(|\langle \mathbf{q}_i,\mathbf{q}_{i+1}\rangle|)$, and translational change by the Euclidean displacement $\lVert \mathbf{t}_{i+1} - \mathbf{t}_i \rVert_2$.
Summing these local changes over the full clip gives the total rotation and translation:
\begin{equation}
R_{\mathrm{tot}} = \sum_{i=1}^{T-1} \Delta\phi_i,
\qquad
D_{\mathrm{tot}} = \sum_{i=1}^{T-1} \lVert \mathbf{t}_{i+1} - \mathbf{t}_i \rVert_2.
\end{equation}
We then choose how many trajectory segments are needed so that each segment covers only a manageable amount of motion.
Concretely, $R_{\mathrm{seg}}$ and $D_{\mathrm{seg}}$ denote the target rotation and translation budget for a single segment.
We compare the total motion of the clip against these budgets along both axes and let the larger normalized motion determine the number of segments, capped at $M_{\max}$:
\begin{equation}
m =
\min\!\left(
M_{\max},
\max\!\left(
1,
\left\lceil
\max\!\left(
\frac{R_{\mathrm{tot}}}{R_{\mathrm{seg}}},
\frac{D_{\mathrm{tot}}}{D_{\mathrm{seg}}}
\right)
\right\rceil
\right)
\right).
\end{equation}
If necessary, we further reduce $m$ so that each segment contains at least $L_{\min}$ frames.
As a result, short or nearly linear trajectories are handled with a single segment, whereas clips with larger accumulated motion are divided into shorter intervals that are easier for the inpainting model to follow.

Once $m$ is fixed, we place the $m-1$ interior keyframes at moments of peak trajectory inflection, the exact points where the generator is most prone to geometric drift.
%
To capture such changes, we compute the local translation and rotation increments, $\mathbf{v}_i = \mathbf{t}_{i+1} - \mathbf{t}_i$ and $\boldsymbol{\omega}_i = \log(\mathbf{q}_{i+1}\mathbf{q}_i^{-1})$, and assign each interior frame a motion-change score:
\begin{equation}
s(i) = \theta(\mathbf{v}_{i-1}, \mathbf{v}_i)
+
\alpha \, \theta(\boldsymbol{\omega}_{i-1}, \boldsymbol{\omega}_i),
\end{equation}
where $\theta(\cdot,\cdot)$ denotes the angle between two vectors, and each term is suppressed when the corresponding motion magnitude is below a small threshold.
A large score indicates a sharp change in translation direction, rotation direction, or both.
We select up to $m-1$ interior keyframes by greedy non-maximum suppression over $s(i)$, enforcing a minimum temporal separation of $L_{\min}$.
If fewer than $m-1$ reliable peaks are selected, we place the remaining keyframes by approximately uniform partition of the largest uncovered gaps.

Given the final keyframe set $K$, we generate each segment independently by conditioning the inpainting model on the boundary frames $x_{k_j}$ and $x_{k_{j+1}}$.
Concatenating the resulting subclips and removing duplicated boundary frames yields the distorted counterpart $\mathbf{y}$.
This adaptive construction keeps the synthesized video close enough to the target that the learning problem remains correction rather than regeneration, while still inducing the geometric failures of interest, such as structural warping, inconsistent parallax, depth-order violations, and shot-like transitions.

\vspace{-2mm}
\subsection{Motion Disentanglement with Optical Flow}
\vspace{-2mm}
Geometry distortions under camera motion arise in part because current video training data contains rich human and object motion, but relatively sparse supervision for viewpoint-induced appearance change.
As a result, video models can confuse camera motion with scene dynamics, and may explain viewpoint change by deforming scene structure instead of preserving a stable 3D layout.
Geometry correction is therefore not only a restoration problem, but also a motion disentanglement problem.

Optical flow provides a natural auxiliary representation for this purpose.
Unlike RGB frames, which entangle appearance and motion, optical flow explicitly represents motion through its location, direction, and magnitude.
Camera motion tends to induce a globally coherent motion field, whereas object motion appears as local deviations from that field.
We therefore compute forward optical flow between consecutive frames of the distorted input video and use it as an additional conditioning signal for the correction model.
Because the flow is computed from the distorted input rather than the target, the same signal is available during both training and inference.
Optical flow does not supervise the model directly; instead, it provides motion structure that helps distinguish what should be corrected from what should be preserved.

\input{fig_text/main_diagram}
\vspace{-2mm}
\subsection{Video Geometry Corrector}
\vspace{-2mm}
\label{sec:model}
Figure~\ref{fig:main_diagram} summarizes the overall data generation and training pipeline.
{\ourmethod} is built on a pretrained video diffusion backbone~\cite{wan} and trained to map a distorted input video $\mathbf{y}$ to its geometry-consistent target $\mathbf{x}$.
Let $E(\cdot)$ denote the video encoder, and let $\mathbf{z} = E(\mathbf{x})$ be the latent representation of the target video.
During training, we sample a diffusion timestep $t$ and Gaussian noise $\epsilon$, and construct the noisy latent $\mathbf{z}_t$ in the standard way.
The denoising network is trained to predict $\epsilon$ conditioned on the distorted input video, the clip text prompt, and the optical flow maps:
\begin{equation}
    \mathcal{L}
    =
    \mathbb{E}_{t,\epsilon}
    \left[
        \left\|
            \epsilon - \epsilon_\theta(\mathbf{z}_t, t, \mathbf{c})
        \right\|_2^2
    \right],
\end{equation}
where $\mathbf{c}$ denotes the distorted-video latent, the text condition, and the optical flow condition.
We inject the video and flow conditions through channel concatenation, while the text prompt is provided through cross-attention.
The model is thus trained to preserve scene content and camera intent while denoising toward a geometry-consistent target.
At inference, {\ourmethod} takes a distorted input video, computes optical flow from it, and performs conditional denoising to produce a geometry-corrected output.
Because the model is trained as a separate video-to-video corrector, it can refine videos from different generators without modifying the source model itself.

%% file: fig_text/keyframes.tex
\begin{figure}[t]
    \centering
    \includegraphics[width=.98\linewidth]{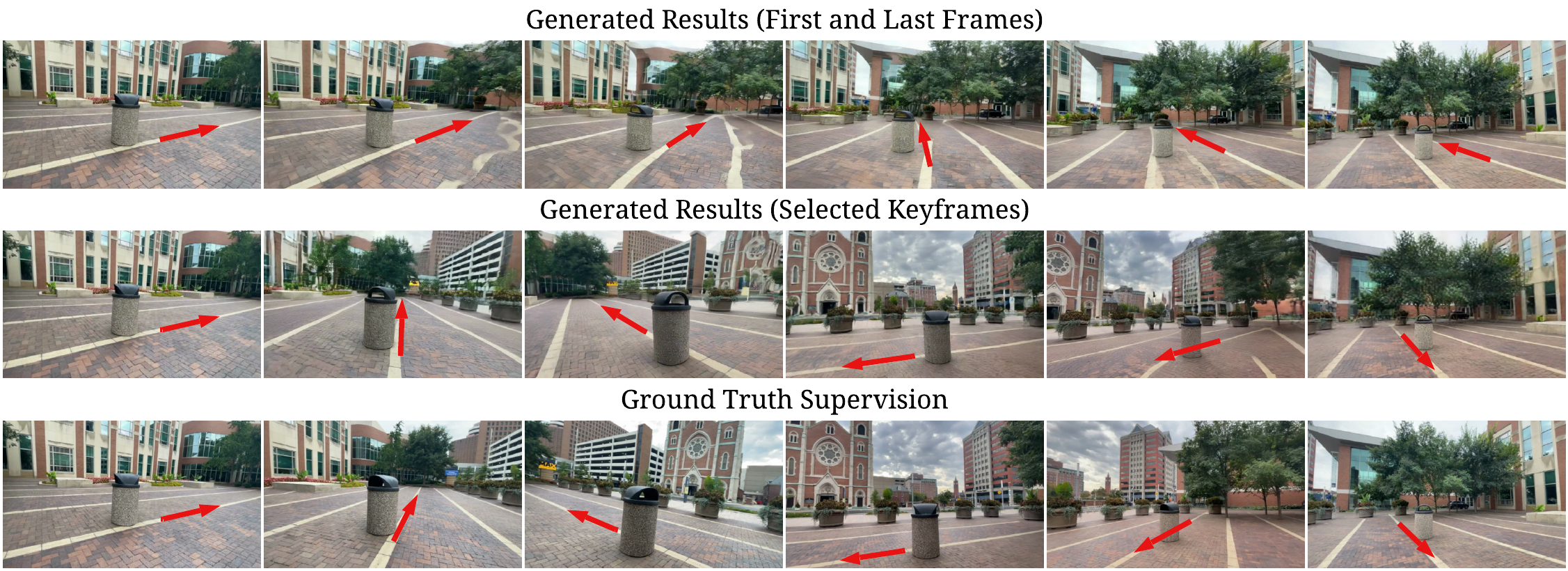}
    \caption{\textbf{Effect of adaptive keyframe selection.} The red arrows indicate a constant 3D direction while the camera is moving. For clips with large, nonlinear camera motion, conditioning the image-to-video model only on the first and last frames often produces a shortcut trajectory that deviates substantially from the ground-truth path (top; about $45^\circ$ rotation versus about $225^\circ$ in the ground truth). Our adaptive keyframe selection inserts intermediate keyframes, generates shorter segments conditioned on their boundary keyframes, and stitches them together. This produces geometry-distorted videos that remain close to the target camera trajectory and are therefore suitable as paired training data (middle), compared with the ground-truth supervision video (bottom).}
    \vspace{-3mm}
    \label{fig:keyframes}
\end{figure}

%% file: fig_text/main_diagram.tex
\begin{figure}[t]
    \centering
    \includegraphics[width=.98\linewidth]{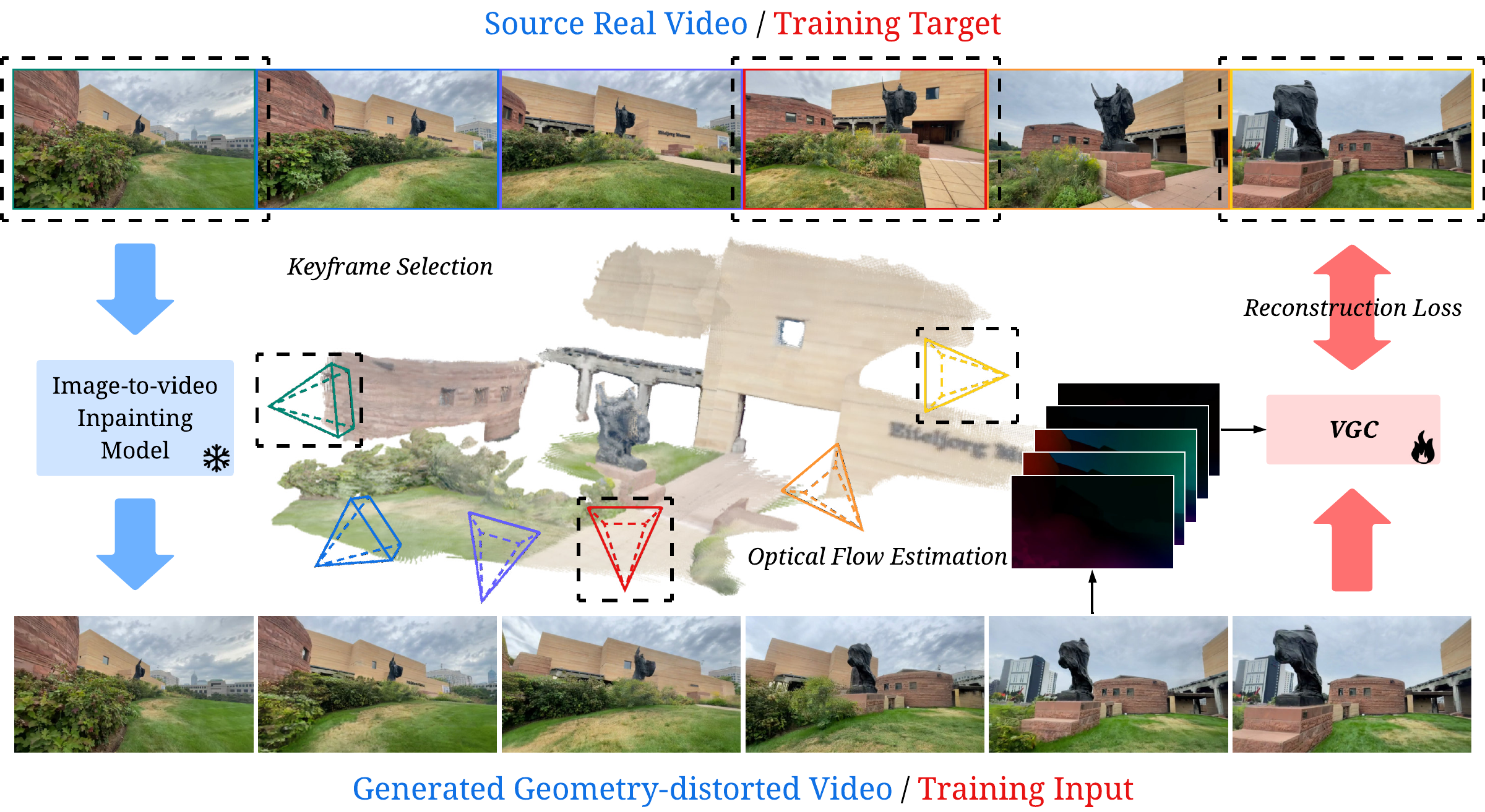}
    \caption{\textbf{Overview of Video Geometry Corrector ({\ourmethod}).} Starting from a real video with diverse camera motion, we select trajectory-adaptive keyframes and use them to condition an image-to-video inpainting model, which produces a geometry-distorted video that serves as the training input. We then train {\ourmethod} to map this distorted input to the geometry-consistent target video with a reconstruction loss. Optical flow is computed from the distorted video and used as an additional conditioning signal for {\ourmethod}, providing motion cues that help disentangle camera motion from scene dynamics. \textit{The 3D reconstruction is shown here only for visualization and is not used in the pipeline.}}
    \vspace{-3mm}
    \label{fig:main_diagram}
\end{figure}

%% file: sec/4_exp.tex
\vspace{-2mm}
\section{Experiments}
\vspace{-2mm}
\subsection{Experimental Setup}
\label{subsec:experimental_setup}
\vspace{-2mm}
\textbf{Implementation details.}
We construct {\ourdataset} from two publicly available datasets, DL3DV~\cite{dl3dv} and SpatialVID~\cite{spatialvid}.
We select 9K videos from DL3DV, reserve the first 1K for evaluation, and filter 63K videos from SpatialVID that contain large camera motion.
To construct paired supervision, we use Wan 2.1-Fun-InP to synthesize geometry-distorted videos from trajectory-adaptive keyframes.
Given a start frame and an end frame, Wan 2.1-Fun-InP generates the intermediate video connecting them.
For clips with complex camera motion, we apply this model independently to multiple keyframe-bounded subsegments and stitch the generated segments together to form the distorted training input.
Because Wan's VAE temporally compresses the video, keyframes are selected so that each generated subsegment has a frame count compatible with the temporal compression schedule.
We use Wan 2.1-1.3B~\cite{wan} as the backbone of {\ourmethod}.
The model is trained for 10K steps on 32 H100 GPUs with AdamW, using a learning rate of $1\times10^{-5}$, weight decay $0.01$, and a global batch size of 64.
Unless otherwise specified, all training videos contain 81 frames at a resolution of $832\times480$.
Optical flow is estimated with SEA-RAFT~\cite{Sea-raft} and used as an additional conditioning signal.
%
%
We use $M_{max}$ = 4, $R_{seg}$ = $120^{\circ}$, $D_{seg}$ = 10 m, $L_{min}$ = 15, and $\alpha$ = 1.

\textbf{Evaluation datasets and metrics.}
To evaluate {\ourmethod}'s performance, we study two scenarios in which camera motion is introduced.
In the first setting, camera motion is specified by camera poses.
Following Geometry Forcing~\cite{geometryforcing}, we evaluate on RealEstate10K~\cite{re10k} using 100 randomly sampled test videos and DFoT~\cite{dfot} as the source generator.
We compare against Geometry Forcing and Geometry Forcing + REPA~\cite{repa}, and report PSNR, SSIM, LPIPS~\cite{lpips}, and revisit error (RVE).
Following~\cite{geometryforcing}, RVE measures long-range consistency under a full camera rotation by comparing the generated view at the end of the trajectory with the corresponding reference view.
In the second setting, camera motion is induced by the text prompt.
We use three source generators in this setting: CogVideoX-5B I2V, CogVideoX-5B T2V, and CogVideoX 1.5-5B T2V.
We randomly sample 200 prompts from the held-out DL3DV test set.
Following VideoGPA~\cite{videogpa}, we reconstruct each generated video independently using Depth Anything 3~\cite{da3} and compare its original frames with frames reprojected from the reconstructed 3D geometry to compute PSNR, SSIM, LPIPS~\cite{lpips}, MSE, and 3D consistency score (3DCS).
Unlike the pose-conditioned setting, these metrics do not require ground-truth video frames.
In both settings, {\ourmethod} is applied as a post-hoc corrector to the outputs of the source generator, demonstrating its flexibility across different video model families.

\input{fig_text/re10k}
\vspace{-2mm}
\subsection{Experimental Results}
\vspace{-2mm}
\textbf{Results on RealEstate10K.}
Table~\ref{tab:gf} shows that {\ourmethod} improves geometric consistency in the pose-conditioned setting and achieves stronger PSNR, SSIM, and RVE than the DFoT baseline~\cite{dfot}.
When applied as a post-hoc corrector to the outputs of Geometry Forcing~\cite{geometryforcing} and Geometry Forcing + REPA~\cite{geometryforcing,repa}, {\ourmethod} further improves their results, demonstrating that it can be flexibly combined with geometry-aware generators.
Figure~\ref{fig:re10k} provides qualitative examples of the improved temporal consistency and scene structure.
Overall, these results support our claim that {\ourmethod} improves both local visual quality and cross-frame geometric consistency under camera motion.

We also conduct an ablation study in this setting to evaluate the effect of optical flow as an auxiliary conditioning signal.
Comparing the full model with the variant without flow in Table~\ref{tab:gf}, we find that optical flow improves PSNR, SSIM, and RVE.
These results suggest that flow helps improve local structural consistency, and that {\ourmethod}'s main gains come from the paired supervision itself.

\input{tab/dl3dv}
\input{fig_text/dl3dv}
\textbf{Results on DL3DV.}
Table~\ref{tab:dl3dv} shows that {\ourmethod} improves reconstruction quality and geometric consistency across all three CogVideoX source generators in both text-to-video and image-to-video settings.
Compared with the original baselines, {\ourmethod} consistently improves PSNR, SSIM, and MSE, indicating better frame fidelity under prompt-driven camera motion.
When applied on top of VideoGPA~\cite{videogpa}, {\ourmethod} further improves its results, showing that the proposed corrector can be flexibly combined with geometry-aware refinement methods.
Figure~\ref{fig:dl3dv} provides qualitative examples in which {\ourmethod} corrects geometric distortions while better preserving scene appearance.

\input{fig_text/wan21}

\input{tab/spatialvid}
\input{fig_text/spatialvid}

\vspace{-2mm}
\subsection{Extension to General Video Refinement}
\vspace{-2mm}
\label{subsec:general_refine}

We further investigate whether the proposed correction framework can extend beyond geometry correction in static scenes.
To this end, we train a larger version of {\ourmethod} based on Wan 2.2-14B and evaluate it in a more general video refinement setting.
With increased model capacity, {\ourmethod} not only corrects geometric distortions under camera motion, but also improves dynamic videos by recovering more plausible human and object geometry.
As shown in Figure~\ref{fig:wan21}, the smaller 1.3B version does not exhibit the same ability to recover dynamic motion.
Because the 14B model takes 480p videos as input and produces 720p outputs, it also performs super-resolution as part of the refinement process.

For evaluation, we randomly sample 100 held-out videos from SpatialVID~\cite{spatialvid} and compare against general video refinement methods, including SeedVR2~\cite{seedvr2} and VEnhancer v2~\cite{venhancer}.
To demonstrate the flexibility of {\ourmethod}, we apply it to videos generated by three different source models: CogVideoX~\cite{congvideox}, HunyuanVideo 1.5~\cite{hunyuanvideo}, and Wan 2.1~\cite{wan}.
CogVideoX and HunyuanVideo 1.5 are evaluated in the image-to-video setting, while Wan 2.1 is evaluated in the inpainting setting.
%
%
Since these videos contain dynamic content and do not admit a geometry-consistent reconstruction target, we evaluate them with no-reference video-quality metrics rather than reconstruction metrics.
Specifically, we report VBench~\cite{vbench} metrics together with DOVER~\cite{dover}, a holistic no-reference video-quality metric that summarizes technical and aesthetic quality.

The quantitative results in Table~\ref{tab:spatialvid} show that {\ourmethod} achieves performance comparable to strong video refinement baselines across a range of perceptual and temporal metrics, despite not being explicitly designed for generic enhancement.
In particular, it remains consistently competitive in subject consistency, background consistency, and motion quality, while achieving strong results in dynamic degree, imaging quality, and DOVER.
These results suggest that, when scaled to larger backbones, the proposed correction framework naturally generalizes to broader video refinement settings.

Figure~\ref{fig:spatialvid} provides qualitative examples.
{\ourmethod} corrects implausible geometry changes, recovers more plausible camera motion, restores distorted structural details, and better preserves human identity and motion continuity.
It also improves local detail through super-resolution while remaining faithful to the source video.

%% file: fig_text/re10k.tex
\begin{center}
\begin{minipage}[c]{0.55\linewidth}
    \centering
    \captionsetup{type=table}
    \caption{\textbf{Quantitative results on RealEstate10K~\cite{re10k}.} {\ourmethod} achieves the strongest PSNR, SSIM, and revisit error (RVE) among the direct baselines. When applied as a post-hoc corrector to Geometry Forcing (GF)~\cite{geometryforcing} and GF + REPA~\cite{geometryforcing,repa}, it further improves their performance, showing that it can be flexibly combined with geometry-aware generators. Comparing {\ourmethod} with and without optical flow shows that flow yields modest gains in image quality while maintaining similar long-range geometric consistency.}
    \label{tab:gf}
    \scriptsize
    \setlength{\tabcolsep}{4pt}
    \resizebox{\linewidth}{!}{
    \begin{tabular}{lcccc}
    \toprule
    Method 
    & PSNR $\uparrow$ & SSIM $\uparrow$ & LPIPS $\downarrow$ & RVE$\downarrow$ \\
    \midrule
    DFoT~\cite{dfot} 
    & 13.61 & 0.4006 & 0.4487 & 184.2 \\
    GF~\cite{geometryforcing} 
    & 14.76 & 0.4484 & \textbf{0.4127} & 181.0 \\
    GF~\cite{geometryforcing} + REPA~\cite{repa}
    & 14.16 & 0.4217 & 0.4218 & 171.8 \\
    DFoT + {\ourmethod} (w/o Flow) 
    & 14.76 & 0.4509 & 0.4433 & 169.4 \\
    DFoT + {\ourmethod} (Ours)
    & \textbf{14.87} & \textbf{0.4545} & 0.4395 & \textbf{168.9} \\
    \midrule
    GF + {\ourmethod} 
    & \metricgain{15.50}{+0.74}
    & \metricgain{0.4715}{+0.0232}
    & \metricloss{0.4150}{+0.0022}
    & \metricgain{158.8}{-22.2} \\
    GF + REPA + {\ourmethod} 
    & \metricgain{15.25}{+1.09}
    & \metricgain{0.4675}{+0.0458}
    & \metricloss{0.4226}{+0.0008}
    & \metricgain{160.9}{-10.9} \\
    \bottomrule
    \end{tabular}
    }
\end{minipage}
\hfill
\begin{minipage}[c]{0.42\linewidth}
    \centering
    \includegraphics[width=\linewidth]{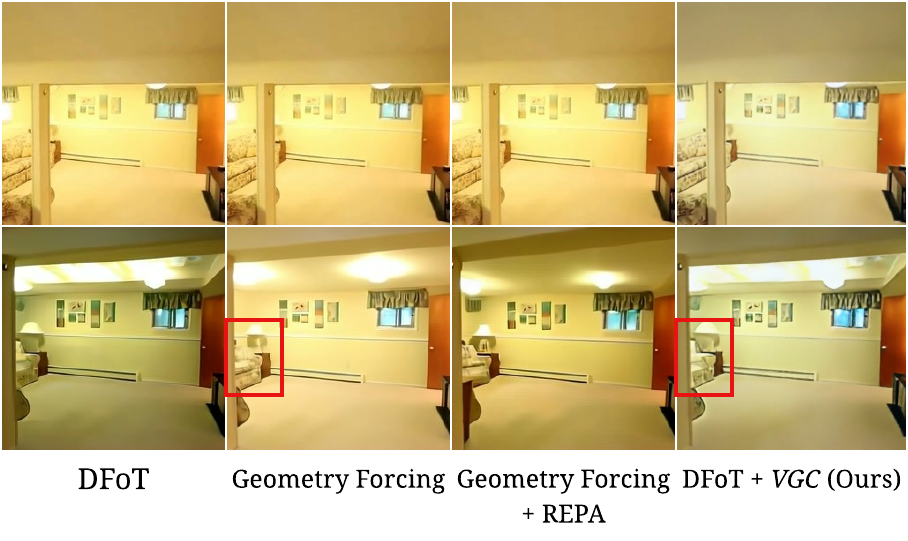}
    \captionsetup{type=figure}
    \vspace{-4mm}
    \caption{\textbf{Visual results on RealEstate10K~\cite{re10k}.} DFoT~\cite{dfot} and GF + REPA~\cite{geometryforcing,repa} produce inconsistent lighting across frames, while GF~\cite{geometryforcing} generates an incomplete sofa structure (red box). {\ourmethod} restores more consistent lighting and more coherent scene geometry.}
    \label{fig:re10k}
\end{minipage}
\end{center}



%% file: tab/dl3dv.tex
\begin{table}[t]
\caption{\textbf{Quantitative results on DL3DV~\cite{dl3dv}.} We evaluate {\ourmethod} on three CogVideoX~\cite{congvideox} source generators in both text-to-video and image-to-video settings. {\ourmethod} consistently improves PSNR, SSIM, and MSE over the corresponding baselines, and further improves VideoGPA~\cite{videogpa} when applied as a post-hoc corrector. These results show that {\ourmethod} transfers across different generation settings and can be flexibly combined with other geometry-aware refinement methods.}
\vspace{2mm}
\label{tab:dl3dv}
\centering
\scriptsize
\setlength{\tabcolsep}{3pt}
\renewcommand{\arraystretch}{1.18}
\resizebox{\linewidth}{!}{
\begin{tabular}{lccccccccccccccc}
\toprule
\multirow{2}{*}{Method}
& \multicolumn{5}{c}{\textit{CogVideoX-5B T2V}}
& \multicolumn{5}{c}{\textit{CogVideoX 1.5-5B T2V}}
& \multicolumn{5}{c}{\textit{CogVideoX-5B I2V}} \\
\cmidrule(lr){2-6} \cmidrule(lr){7-11} \cmidrule(lr){12-16}
& PSNR$\uparrow$ & SSIM$\uparrow$ & LPIPS$\downarrow$ & MSE$\downarrow$ & 3DCS$\downarrow$
& PSNR$\uparrow$ & SSIM$\uparrow$ & LPIPS$\downarrow$ & MSE$\downarrow$ & 3DCS$\downarrow$
& PSNR$\uparrow$ & SSIM$\uparrow$ & LPIPS$\downarrow$ & MSE$\downarrow$ & 3DCS$\downarrow$ \\

\midrule

Baseline
& 15.6668 & 0.5389 & 0.5285 & 0.0333 & 0.8617
& 15.9062 & 0.4644 & 0.5449 & 0.0289 & 0.8342
& 17.3189 & 0.5543 & 0.5760 & 0.0218 & 0.7944 \\

Baseline + {\ourmethod}
& \makecell{16.6940 \\ {\scriptsize\textcolor{darkgreen}{(+1.0272)}}}
& \makecell{0.5519 \\ {\scriptsize\textcolor{darkgreen}{(+0.0130)}}}
& \makecell{0.5733 \\ {\scriptsize\textcolor{darkred}{(+0.0448)}}}
& \makecell{0.0264 \\ {\scriptsize\textcolor{darkgreen}{(-0.0069)}}}
& \makecell{0.8376 \\ {\scriptsize\textcolor{darkgreen}{(-0.0241)}}}
& \makecell{17.7342 \\ {\scriptsize\textcolor{darkgreen}{(+1.8280)}}}
& \makecell{0.5284 \\ {\scriptsize\textcolor{darkgreen}{(+0.0640)}}}
& \makecell{0.5936 \\ {\scriptsize\textcolor{darkred}{(+0.0487)}}}
& \makecell{0.0192 \\ {\scriptsize\textcolor{darkgreen}{(-0.0097)}}}
& \makecell{0.7859 \\ {\scriptsize\textcolor{darkgreen}{(-0.0483)}}}
& \makecell{17.9989 \\ {\scriptsize\textcolor{darkgreen}{(+0.6800)}}}
& \makecell{0.5456 \\ {\scriptsize\textcolor{darkred}{(-0.0087)}}}
& \makecell{0.6284 \\ {\scriptsize\textcolor{darkred}{(+0.0524)}}}
& \makecell{0.0184 \\ {\scriptsize\textcolor{darkgreen}{(-0.0034)}}}
& \makecell{0.8121 \\ {\scriptsize\textcolor{darkred}{(+0.0177)}}} \\

\midrule

VideoGPA~\cite{videogpa}
& 14.6318 & 0.5210 & 0.5065 & 0.0428 & 0.9346
& 14.4002 & 0.4175 & 0.5531 & 0.0408 & 0.9614
& 15.7203 & 0.5240 & 0.5749 & 0.0320 & 0.8952 \\

VideoGPA~\cite{videogpa} + {\ourmethod}
& \makecell{15.9092 \\ {\scriptsize\textcolor{darkgreen}{(+1.2774)}}}
& \makecell{0.5413 \\ {\scriptsize\textcolor{darkgreen}{(+0.0203)}}}
& \makecell{0.5546 \\ {\scriptsize\textcolor{darkred}{(+0.0481)}}}
& \makecell{0.0319 \\ {\scriptsize\textcolor{darkgreen}{(-0.0109)}}}
& \makecell{0.8737 \\ {\scriptsize\textcolor{darkgreen}{(-0.0609)}}}
& \makecell{16.8430 \\ {\scriptsize\textcolor{darkgreen}{(+2.4428)}}}
& \makecell{0.5039 \\ {\scriptsize\textcolor{darkgreen}{(+0.0864)}}}
& \makecell{0.5907 \\ {\scriptsize\textcolor{darkred}{(+0.0376)}}}
& \makecell{0.0235 \\ {\scriptsize\textcolor{darkgreen}{(-0.0173)}}}
& \makecell{0.8258 \\ {\scriptsize\textcolor{darkgreen}{(-0.1356)}}}
& \makecell{17.3053 \\ {\scriptsize\textcolor{darkgreen}{(+1.5850)}}}
& \makecell{0.5379 \\ {\scriptsize\textcolor{darkgreen}{(+0.0139)}}}
& \makecell{0.6216 \\ {\scriptsize\textcolor{darkred}{(+0.0467)}}}
& \makecell{0.0225 \\ {\scriptsize\textcolor{darkgreen}{(-0.0095)}}}
& \makecell{0.8463 \\ {\scriptsize\textcolor{darkgreen}{(-0.0489)}}} \\
\bottomrule
\vspace{-2mm}
\end{tabular}
}
\end{table}

%% file: fig_text/dl3dv.tex
\begin{figure}[t]
    \centering
    \includegraphics[width=.8\linewidth]{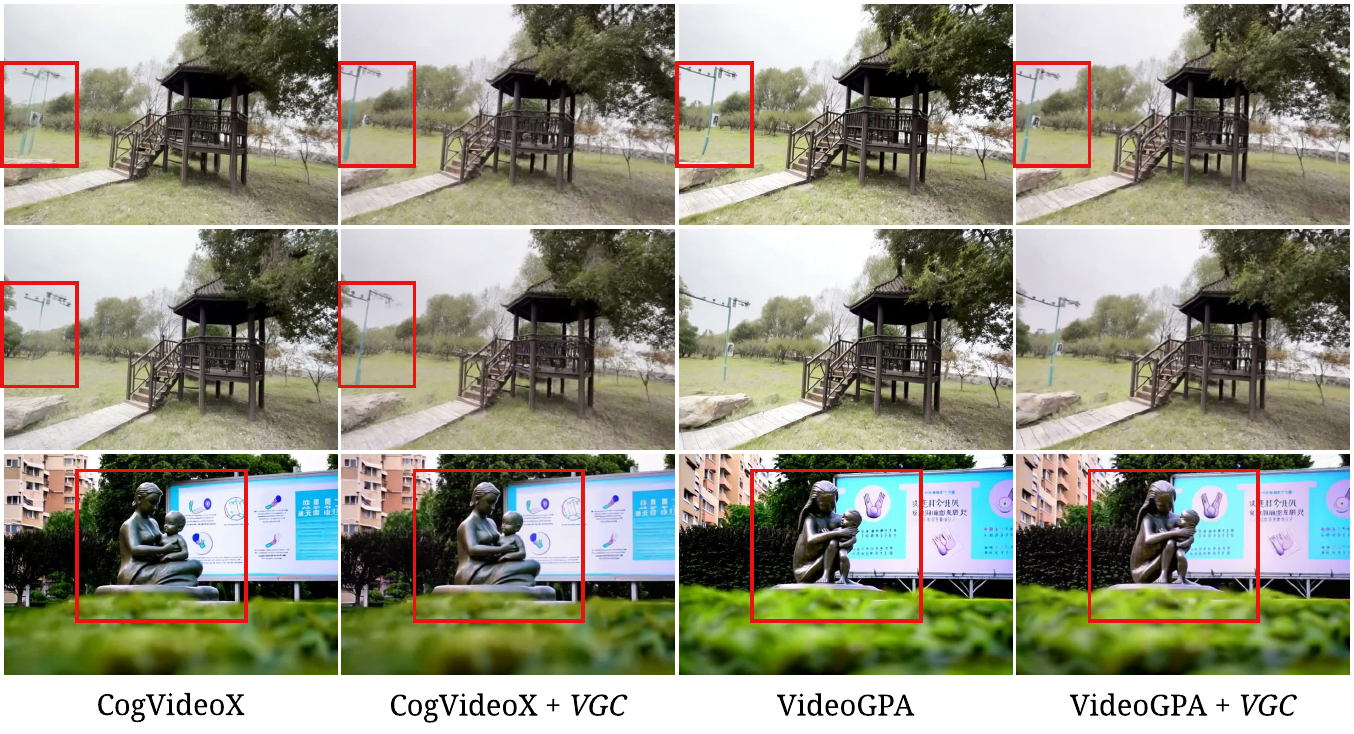}
    \caption{\textbf{Visual results on DL3DV~\cite{dl3dv}.} The first two rows show image-to-video results. In the baseline CogVideoX~\cite{congvideox} outputs, the streetlight is duplicated, while VideoGPA~\cite{videogpa} produces a distorted and tilted streetlight (red boxes). {\ourmethod} corrects these geometric errors and restores a more plausible structure. The last row shows a text-to-video example, where {\ourmethod} better preserves the original appearance of the sculpture, while VideoGPA changes its structure and other scene details.}
    \vspace{-1mm}
    \label{fig:dl3dv}
\end{figure}

%% file: fig_text/wan21.tex
\begin{figure}[t]
    \centering
    \includegraphics[width=.98\linewidth]{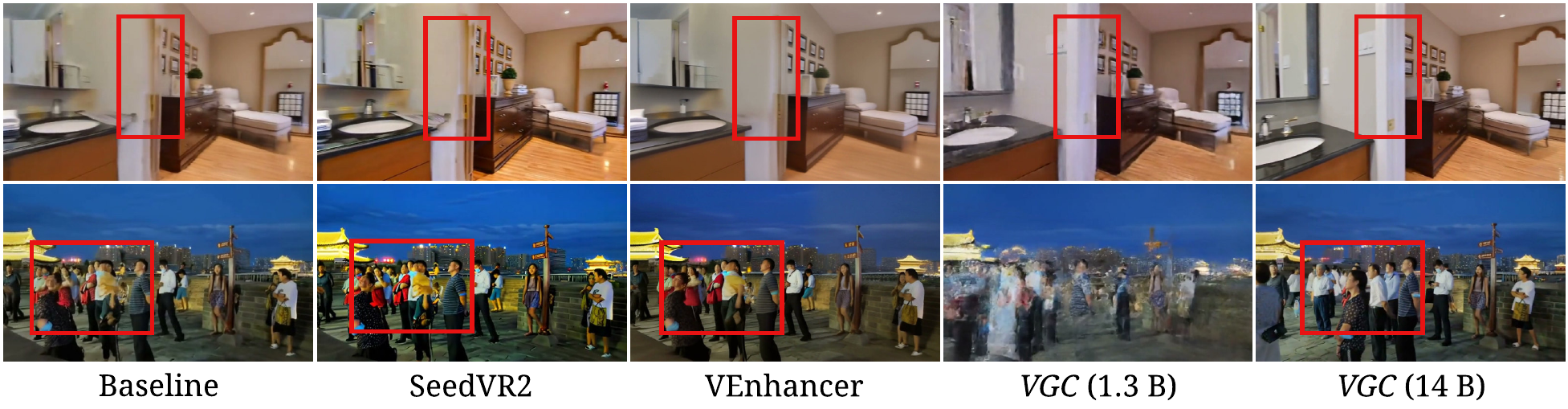}
    \caption{\textbf{Comparison of model sizes.} In static scenes (top), both {\ourmethod} 1.3B and 14B correct geometric distortions produced by the baseline model that SeedVR2~\cite{seedvr2} and VEnhancer~\cite{venhancer} fail to restore (red boxes). The 14B model further improves the geometry of moving humans and objects in dynamic scenes (bottom).}
    \vspace{-3mm}
    \label{fig:wan21}
\end{figure}

%% file: tab/spatialvid.tex
\begin{table}[t]
\centering
\caption{\textbf{General video refinement results on SpatialVID~\cite{spatialvid}.} We compare {\ourmethod} with general video refinement and enhancement methods~\cite{seedvr2, venhancer} on dynamic videos generated by CogVideoX~\cite{congvideox}, HunyuanVideo~\cite{hunyuanvideo}, and Wan~\cite{wan}. Although {\ourmethod} is designed for geometry correction under camera motion rather than generic enhancement, it achieves competitive performance across multiple perceptual and temporal metrics, and is particularly strong in dynamic degree, imaging quality, and DOVER~\cite{dover}. These results suggest that the proposed correction framework extends beyond static-scene geometry and remains effective in more general settings.}
\vspace{2mm}
\label{tab:spatialvid}
\scriptsize
\setlength{\tabcolsep}{4pt}
\renewcommand{\arraystretch}{1.12}
\resizebox{\linewidth}{!}{
\begin{tabular}{lcccccccc}
\toprule
\multirow{2}{*}{Method}
& \makecell{Subject\\Consistency}
& \makecell{Background\\Consistency}
& \makecell{Temporal\\Flickering}
& \makecell{Motion\\Smoothness}
& \makecell{Dynamic\\Degree}
& \makecell{Aesthetic\\Quality}
& \makecell{Imaging\\Quality}
& DOVER~\cite{dover} \\
\midrule

\multicolumn{9}{c}{\textit{CogVideoX-I2V~\cite{congvideox}}} \\
Baseline
& 0.9480 & 0.9455 & 0.9742 & \underline{0.9918} & 0.30 & 0.5095 & 69.6945 & 9.54 \\
SeedVR2~\cite{seedvr2}
& 0.9447 & \underline{0.9476} & 0.9603 & 0.9829 & 0.30 & \textbf{0.5313} & \underline{71.1767} & 8.31 \\
VEnhancer v2~\cite{venhancer}
& \textbf{0.9506} & \textbf{0.9602} & \underline{0.9749} & \textbf{0.9920} & \underline{0.31} & \underline{0.5237} & 69.3742 & \underline{10.31} \\
{\ourmethod} (Ours)
& \underline{0.9483} & 0.9450 & \textbf{0.9772} & 0.9891 & \textbf{0.66} & 0.5011 & \textbf{72.7654} & \textbf{10.50} \\

\midrule
\multicolumn{9}{c}{\textit{HunyuanVideo 1.5-I2V~\cite{hunyuanvideo}}} \\
Baseline
& 0.8970 & 0.9231 & \underline{0.9592} & \underline{0.9829} & \textbf{1.00} & 0.5133 & \underline{73.7167} & \textbf{10.10} \\
SeedVR2~\cite{seedvr2}
& 0.8967 & 0.9167 & 0.9415 & 0.9714 & \textbf{1.00} & \textbf{0.5539} & 72.2762 & 7.89 \\
VEnhancer v2~\cite{venhancer}
& \textbf{0.9061} & \textbf{0.9294} & \textbf{0.9596} & \textbf{0.9860} & \underline{0.99} & \underline{0.5301} & 71.9688 & \underline{9.85} \\
{\ourmethod} (Ours)
& \underline{0.8995} & \underline{0.9282} & 0.9482 & 0.9799 & 0.96 & 0.4849 & \textbf{75.1367} & 9.30 \\

\midrule
\multicolumn{9}{c}{\textit{Wan 2.1-Fun-InP~\cite{wan}}} \\
Baseline
& \underline{0.9247} & 0.9239 & \textbf{0.9712} & \textbf{0.9914} & \underline{0.87} & 0.5043 & 69.5165 & \underline{10.87} \\
SeedVR2~\cite{seedvr2}
& 0.9217 & 0.9231 & 0.9547 & 0.9809 & \underline{0.87} & \textbf{0.5431} & \textbf{72.9463} & 8.23 \\
VEnhancer v2~\cite{venhancer}
& 0.9242 & \textbf{0.9371} & \underline{0.9702} & \underline{0.9909} & 0.85 & \underline{0.5254} & 70.6322 & 10.78 \\
{\ourmethod} (Ours)
& \textbf{0.9268} & \underline{0.9281} & 0.9641 & 0.9907 & \textbf{0.89} & 0.5141 & \underline{72.6499} & \textbf{11.47} \\
\bottomrule
\vspace{-3mm}
\end{tabular}
}
\end{table}

%% file: fig_text/spatialvid.tex
\begin{figure}[t]
    \centering
    \includegraphics[width=.98\linewidth]{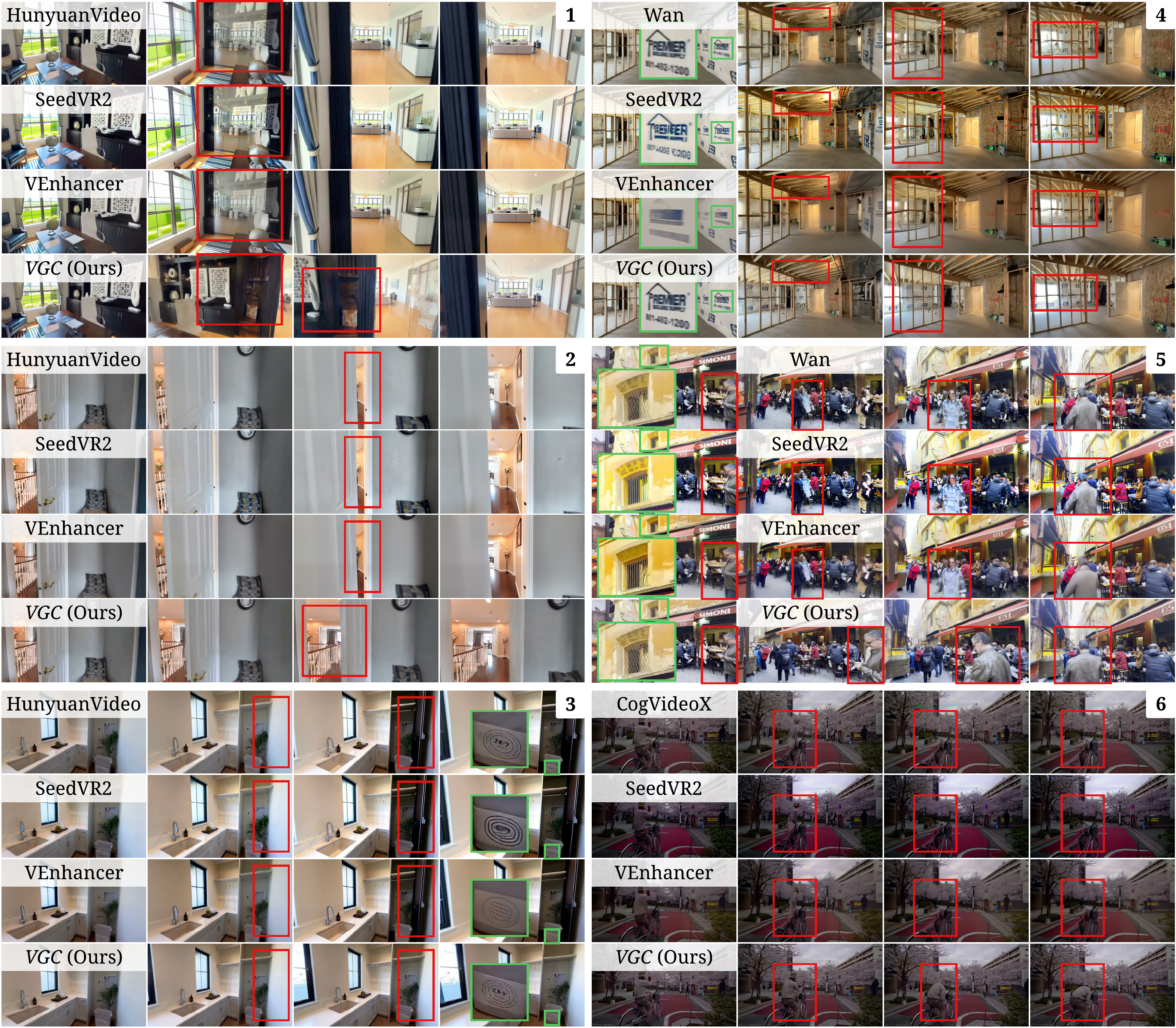}
    \caption{\textbf{General video refinement results.} {\ourmethod} corrects several geometric failures produced by the baseline models. In scene 1, HunyuanVideo~\cite{hunyuanvideo} changes the cabinet into a glass door, while {\ourmethod} restores the correct scene structure and camera motion, placing the door on the right side of the cabinet. In scene 2, HunyuanVideo generates a camera trajectory that passes through the wrong side of the doorway, whereas {\ourmethod} recovers a plausible path. In scenes 3 and 4, {\ourmethod} restores a consistent appearance and recovers fine geometric structure, such as the wooden supports. In the dynamic scenes, {\ourmethod} better preserves identity and motion continuity: in scene 5, it maintains the same man in the leather jacket across frames and recovers a plausible motion trajectory, whereas the Wan baseline fails to infer coherent human motion; in scene 6, it restores the missing cyclist produced by CogVideoX~\cite{congvideox}. The green boxes further show that {\ourmethod} enhances local detail while remaining more faithful to the source video than SeedVR2~\cite{seedvr2} and VEnhancer~\cite{venhancer}. \textit{Please zoom in for details.}}
    \vspace{-3mm}
    \label{fig:spatialvid}
\end{figure}


%% file: sec/5_conclusion.tex
\vspace{-3mm}
\section{Conclusion}
\vspace{-3mm}
We presented Video Geometry Corrector ({\ourmethod}), a post-hoc video-to-video model for correcting geometric distortions in generated videos under camera motion.
We argued that these failures are largely a data problem: current video training data provides limited supervision for viewpoint-induced appearance change, causing camera motion to be entangled with scene dynamics.
To address this, we constructed {\ourdataset} from real videos with diverse camera motion and trained {\ourmethod} to map geometry-distorted videos to geometry-consistent targets.
We further incorporated optical flow as an auxiliary cue to help distinguish what should be corrected from what should be preserved.
Experiments on pose-conditioned and prompt-driven benchmarks show that {\ourmethod} improves geometric consistency across different source generators and can be flexibly combined with existing geometry-aware methods.
With increased model capacity, the same correction framework also extends beyond static-scene geometry toward more general video refinement.

%% file: sec/7_ack.tex
\vspace{-3mm}
\section*{Acknowledgments}
\vspace{-3mm}
This work was supported in part by the Institute of Information \& Communications Technology Planning \& Evaluation (IITP) grant funded by the Korean Government (MSIT) (No. RS-2024-00457882, National AI Research Lab Project) and SONY.